\documentclass[cameraready]{Interspeech}

\title{Beyond WER: Entity and Disfluency Recall in Accented Conversational ASR}

\author[affiliation={1}, orcid=0000-0002-9089-994X]{Fiza}{Husain}
\author[affiliation={1}, orcid=0009-0009-1824-4752]{Ankit}{Pandey}
\author[affiliation={1}, orcid=0009-0006-7587-3991]{Yash}{Singh}

\address{
    $^1$ Stimuler, Bengaluru, India
}

\email{fiza@stimuler.tech, ankit@stimuler.tech, yash@stimuler.tech}


\newcommand{\red}[1]{\textcolor{red}{#1}}
\usepackage{comment}
\usepackage{multirow}

\begin{document}

\maketitle

\begin{abstract}
ASR systems optimised for Word Error Rate (WER) often miss named entities and filled pauses in accented conversational English, both critical for language-learning feedback. We present a three-stage pipeline for speakers from India, Indonesia, and Latin America: (1) heuristic SQL filters curating entity-rich training data at ${\sim}2.8\times$ the entity density of random sampling, (2)~regional LoRA adapters fine-tuned on
Qwen2.5-Omni-3B producing both verbatim and corrected transcripts in a single forward pass, and (3)~a six-category error taxonomy validated by an LLM-based judge (83.8\% agreement, 210~human-labelled samples).  The pipeline achieves 80--85\% entity recall (up from 53--55\%), 76--86\% filler recall (up from $<$5\%), and 6--10\% WER across 6\,k test utterances, outperforming Whisper and a commercial ASR on entity recall while matching a zero-shot 30B model with 10$\times$ fewer parameters. Paired bootstrap tests confirm that curation alone accounts for 2.8--4.2 pp of entity recall gain ($p<0.0001$).
\end{abstract}

\section{Introduction}
\label{sec:intro}

Speech is fundamental to human communication and for English language learners practicing conversational fluency, accurate speech recognition is not merely about minimising Word Error Rate (WER) or Character Error Rate (CER); it is about sustaining trust in the learning loop.  In conversational language-learning platforms, ASR must faithfully capture what a learner actually said---including disfluencies, cultural references, and regionally specific entities---so that meaningful feedback can follow.  In practice, learners need two complementary views of their speech: (i)~a \textbf{verbatim transcript} that retains fillers such as \emph{uh} and \emph{um} for fluency self-assessment, and (ii)~a \textbf{corrected transcript} in which entities are properly spelled and minor structural errors are smoothed, enabling comprehension-focused feedback.  Commercial ASR systems, however, are optimised for aggregate WER and provide no mechanism to control this trade-off.

Evaluating several widely used ASR systems on conversational English from India, Indonesia, and Latin America exposes a stark mismatch between headline WER and downstream utility.  Baseline systems achieve 13--21\% WER---figures that might appear acceptable---yet entity recall ranges from only 53--55\%, and filler recall is near zero ($<$5\%). The disconnect arises because WER penalises a harmless tense shift ("was" → "is") and a garbled \emph{``Agatha Christie''} $\rightarrow$ \emph{``agatha cristee''} equally, despite vastly different pedagogical consequences.

Recent work has introduced entity-focused ASR benchmarks~\cite{del2021earnings} and demonstrated that entity errors propagate irrecoverably to downstream NER systems~\cite{szymanski2023aren}.  Post-hoc LLM correction pipelines attempt to mitigate these errors: Chen et al.~\cite{chen2023hyporadise} use N-best hypothesis re-ranking, while Pusateri et al.~\cite{pusateri2025retrieval} retrieve entity candidates from a vector database.  These approaches leave the upstream acoustic model unchanged and add inference-time complexity. A complementary line applies LoRA~\cite{hu2022lora} to adapt ASR models for accented speech~\cite{bagat2025mixture}, yet existing studies sample training data randomly without regard for entity density.

We do not propose a new architecture or training objective.  Instead, we validate that a simple, reproducible, three-stage pipeline yields large, statistically significant gains on entity recall and filler preservation across three diverse speaker populations.  Our contributions are:
\begin{enumerate}[leftmargin=*,itemsep=2pt,topsep=0pt]
  \item \textbf{Heuristic entity-rich data curation.}\enspace
    Lexical filters applied to existing transcripts surface utterances with ${\sim}2.8\times$ higher entity density than random sampling. Paired bootstrap tests confirm that models trained on this curated data improve entity recall by 2.8--4.2~percentage points over random selection ($p < 0.0001$ across all regions), establishing data curation as a significant and cost-effective lever for entity-focused ASR.

  \item \textbf{Regional LoRA adaptation with dual-output generation.}\enspace
    Fine-tuning Qwen2.5-Omni-3B with standard rank-32 LoRA on 10\,k curated
    samples per region yields 80--85\% entity recall (up from 53--55\%) and
    76--86\% filler recall (up from $<$5\%), while reducing WER to
    6--10\% - a 53--60\% relative reduction.  The model produces both verbatim
    and corrected transcripts in a single forward pass.

  \item \textbf{A diagnostic error taxonomy.}\enspace
    We introduce a six-category error framework - spanning phonetic confusion,
    hallucination, acoustic out-of-vocabulary errors, named-entity
    misspellings, non-actionable audio, and incomplete transcription - validated
    via an LLM-based judge which enables systematic diagnosis of residual errors.
\end{enumerate}

Our results suggest that for accented
conversational speech, \emph{what} you train on matters at least as
much as \emph{how} you train -- and that lightweight curation is a practical first step toward entity-aware ASR without the inference-time cost of post-hoc LLM correction.

\section{Related Work}
\label{sec:related}

\subsection{Parameter-Efficient Fine-Tuning for ASR}

Large pre-trained models such as Whisper~\cite{radford2023robust} and
Wav2Vec2~\cite{baevski2020wav2vec} have shifted focus to efficient
adaptation.  LoRA~\cite{hu2022lora} injects trainable low-rank matrices
into frozen Transformer layers; applied to Whisper for multilingual
ASR~\cite{song2024lora} and code-switching~\cite{yang2025adapting},
rank-16 to rank-64 adapters match full fine-tuning at a fraction of the
parameters.  Memory-efficient implementations such as
Unsloth~\cite{unsloth2023} further lower the hardware barrier.  However,
none of these studies target entity recall as a training objective.  We
pair standard LoRA (rank-32) with a data curation step that enriches the
training set for entity density.

\subsection{Entity-Aware Speech Recognition}

Named entity recognition from speech has evolved from broadcast-news
systems~\cite{kubala1998named} to analyses of ASR-NER error propagation:
Szyma\'{n}ski et al.~\cite{szymanski2023aren} showed that entity errors
propagate irrecoverably to downstream NER taggers.
WhisperNER~\cite{ayache2024whisperner} jointly optimises transcription
and entity tagging but requires entity-type prompts at inference.
Contextual biasing injects entity lists during
decoding~\cite{williams2018contextual}; retrieval-augmented
post-processing corrects names via an
LLM~\cite{pusateri2025retrieval}; and Ling and
Ye~\cite{ling2025customizing} use LLM feedback as a reward signal for
RL-based ASR fine-tuning.  All add inference-time complexity or require
an auxiliary LLM, and none recover from acoustic-level failures.
Liang et al.~\cite{liang2023improving} augment entity-rich data via
speech editing but lack the prosodic variation of real conversational
audio.  Our data-centric approach reduces entity errors at training time
without entity-type labels or an auxiliary LLM at serving time.

\subsection{ASR for Accented and Non-Native Speech}

Multilingual models such as Whisper and
SeamlessM4T~\cite{barrault2023seamlessm4t} have advanced accented-speech
recognition, yet evaluations remain dominated by aggregate WER.  Graham
and Roll~\cite{graham2024evaluating} document persistent WER gaps for
non-native accents.  Bagat et al.~\cite{bagat2025mixture} propose
accent-specific LoRA experts on L2-ARCTIC~\cite{zhao2018l2},
showing that per-accent adapters outperform standard LoRA and full
fine-tuning---motivating our per-region design. Few studies report entity recall, despite entity words carrying the highest
communicative load in learner utterances.

\subsection{Disfluency Modelling in ASR}

Filled pauses constitute 5--10\% of spontaneous
speech~\cite{shriberg2001errrr} and serve as communicative
signals~\cite{clark2002using}, yet Amann et al.~\cite{amann2024augmenting}
show Whisper correctly transcribes only 56\% of disfluent words.  L2
learners produce fillers at more than twice the native
rate~\cite{gotz2013fluency}, making filler frequency a key fluency
indicator~\cite{tavakoli2025assessment}.  Recent work recovers fillers
via modified CTC forced alignment~\cite{amann2024augmenting} or
domain-specific fine-tuning~\cite{akinrintoyo2025whisperd}, but neither
targets L2 speech nor pairs filler preservation with entity-aware
transcription.

\section{Methodology}
\label{sec:method}

\begin{figure*}[t]
  \centering
  \includegraphics[width=\textwidth]{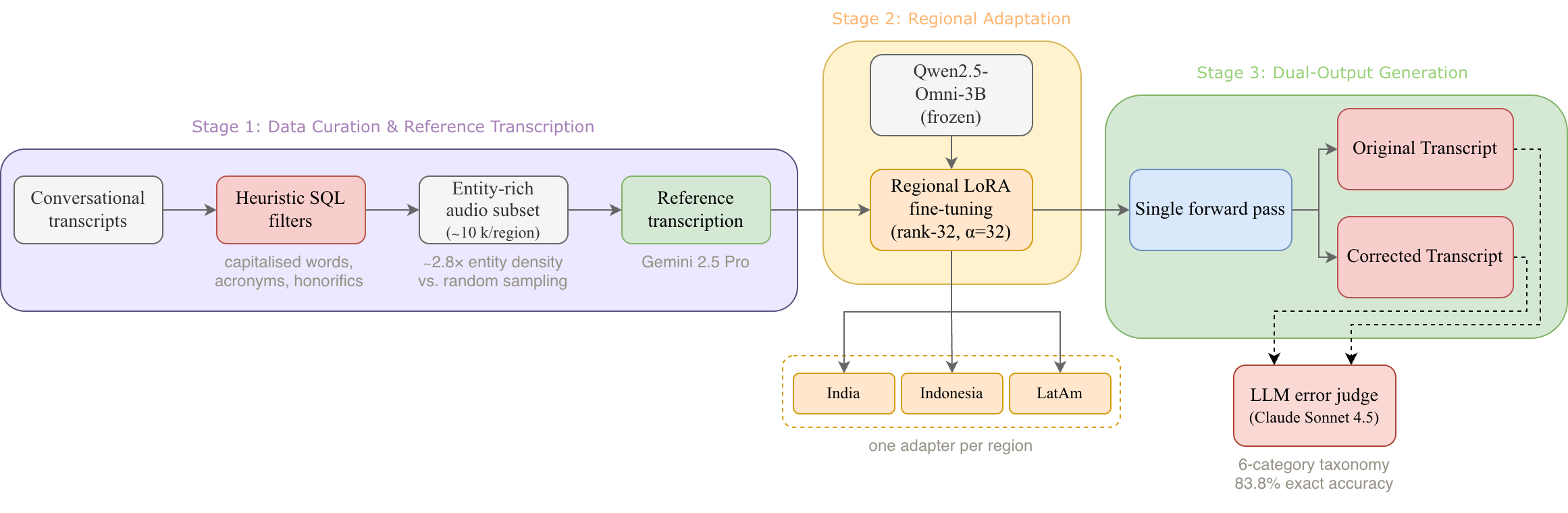}
  \vspace{4pt}
  \caption{Overview of the proposed three-stage pipeline: Stage 1: Heuristic SQL filters select entity-rich utterances (${\sim}2.8\\times$ enrichment); reference transcripts are generated by Gemini 2.5 Pro (750 human-verified samples). Stage 2: Per-region LoRA adapters are trained on Qwen2.5-Omni-3B. Stage 3: A single forward pass produces both a verbatim transcript and a corrected transcript. Dashed arrows show the diagnostic evaluation path via the LLM-based error judge.}
  \label{fig:pipeline}
\end{figure*}

We describe a three-stage pipeline for adapting a multimodal speech
model to accented, entity-rich conversational English.  Each stage uses
standard techniques; the contribution lies in their systematic
combination and empirical validation.  Figure~\ref{fig:pipeline}
provides an overview.

\subsection{Heuristic Entity-Rich Data Curation}
\label{sec:curation}

Randomly sampled conversational audio is entity-sparse: only 24--28\% of utterances in our production logs contain at least one proper noun. To concentrate supervision on the tokens where ASR systems fail most consequentially, we apply lightweight lexical heuristics to existing transcripts stored in a BigQuery data warehouse\footnote{Training data is drawn from proprietary production logs and cannot
be released. Training code will be made publicly available.}.  Specifically, we retain utterances that satisfy \emph{any} of the following SQL-level filters applied to the corrected transcript field: (a)~\textbf{Consecutive capitalised words} — matches sequences such as \emph{Taylor Swift} or \emph{Machu Picchu}, (b)~\textbf{Acronyms} — matches tokens of two or more uppercase letters such as \emph{MBA} or \emph{UNESCO}, (c)~\textbf{Honorifics and titles} — matches prefixes such as \emph{Mr, Dr, Prof, President,} etc, followed by a space, or (d)~\textbf{Mid-sentence capitalisation} — matches capitalised words that do not follow sentence-ending punctuation, capturing entity mentions embedded in running speech.  
All filters require a minimum transcript length of four
words.  This curation yields datasets with 70--78\% entity density
(${\sim}2.8\times$ the random baseline), reducing annotation effort by
${\sim}$65\%.  We apply filters independently per region and sample
10\,k utterances per region from the filtered pool.

\subsection{Reference Transcription}
\label{sec:reference}

Manual inspection of disagreements between professional annotators and
model predictions revealed that Gemini~2.5~Pro~\cite{comanici2025gemini}
produced transcripts closer to the intended speech---particularly for
regionally specific entities (local food names, band names, cultural references) where annotators unfamiliar with the speaker's cultural context introduced errors.  We therefore adopted
Gemini~2.5~Pro as the reference transcriber, prompting it with raw
audio.  Within each region, 250~utterances (750 total) were
independently verified by a human annotator as a gold-standard
validation subset.

\subsection{Dual-Output Training Format}
\label{sec:dualoutput}

The model produces two outputs in a single forward pass via a
structured JSON response: an \textbf{original transcript} that preserves filled pauses from a closed set (\emph{um}, \emph{uh}, \emph{ah}, \emph{er}, \emph{hm}) at their spoken positions, accent-influenced pronunciations, and false starts; and a \textbf{corrected transcript} that replaces mispronounced words with their intended forms, resolves entity spelling, and removes stuttering-induced repetitions while retaining the speaker's original content words and sentence structure. A boolean comprehensibility flag gates both outputs:
when the audio is too degraded, the model returns empty strings rather
than hallucinating.

This multi-task formulation serves two purposes.  First, the original
transcript supports fluency self-assessment while the corrected version
feeds the downstream dialogue system.  Second, jointly learning
corrected entity spellings provides an auxiliary training signal that
improves entity recognition in the original transcript through shared
acoustic representations.  \textbf{All metrics reported in this paper
are computed on the original (verbatim) transcript}, which is the
harder evaluation target.




\subsection{Regional LoRA Adaptation}
\label{sec:lora}

We fine-tune Qwen2.5-Omni-3B~\cite{xu2025qwen2} separately for each region using LoRA~\cite{hu2022lora}. Training a dedicated adapter per region allows the model to specialise in the phonetic, prosodic, and lexical characteristics of each speaker population (e.g., retroflex consonants in Indian English, vowel shifts in Latin American English) without cross-region interference.

\noindent\textbf{LoRA configuration.}
We use rank $r = 32$ and scaling factor $\alpha = 32$, applied to the attention projection matrices of the LLM component. All original model weights are frozen; only the low-rank matrices are updated.

\noindent\textbf{Training details.}
Each regional adapter is trained for 2 epochs on 9\,k utterances (90\% of the 10\,k curated set; the remaining 10\% is held out for validation).  We use the AdamW-8bit optimiser~\cite{dettmers20218} with learning rate $5 \times 10^{-5}$, cosine scheduling, 10\% warmup ratio, weight decay of 0.01, per-device batch size of 4, and maximum sequence length of 2\,048 tokens.  Training is performed using the SFTTrainer framework with Unsloth's~\cite{unsloth2023} memory-efficiency patches (gradient checkpointing, optimised embedding handling), enabling fine-tuning on a single GPU. Best checkpoints are selected by validation loss, evaluated every 300 steps. Each regional adapter trains in approximately 6~hours (2~epochs) on a single NVIDIA A100-40GB GPU. At inference time, adapters are served via vLLM \cite{kwon2023efficient}, achieving a P95 latency of ${\sim}800$ ms.
\noindent\textbf{All-region adapter.}
In addition to the three regional adapters, we train a fourth \emph{all-geo} adapter on a geographically balanced mixture of data from remaining regions, targeting speakers outside the three primary regions.

\subsection{Error Taxonomy and LLM-Based Judge}
\label{sec:taxonomy}
To diagnose residual errors and guide future data collection, we develop a six-category taxonomy:
\textbf{C1}~phonetic substitution (e.g., \emph{tea break}~$\to$~\emph{the break}),
\textbf{C2}~named entity error (any proper noun misspelled or garbled; takes priority),
\textbf{C3}~acoustic gibberish/OOV,
\textbf{C4}~hallucination (inserted content not in audio),
\textbf{C5}~omission (missing content words), and
\textbf{C6}~non-actionable audio (too degraded for any transcriber).
C1--C4 represent actionable errors; C5--C6 are potentially
unrecoverable.

We operationalise this as an LLM-based judge that receives a reference
and ASR transcript and returns a structured JSON with the primary error
category.  To select the judge, we benchmarked 18~LLMs against a
210-sample human-labelled set stratified across regions.
Claude~Sonnet~4.5~\cite{anthropic2025sonnet45} achieved the highest
agreement: 83.8\% exact accuracy, with F1 of 93.1\% on entity errors
($n{=}63$) and 85.2\% on phonetic substitutions ($n{=}65$).  The judge
is used as a \emph{diagnostic tool}; core quantitative claims do not
depend on its accuracy.

\subsection{Evaluation Protocol}
\label{sec:eval}

\noindent\textbf{Test Set.} Each regional adapter is evaluated on a held-out test set of
${\sim}$2\,k utterances that is fully non-overlapping with the
10\,k training pool.  Test utterances are drawn from the same
production distribution and time period but are never seen during
training or validation.

\noindent\textbf{Metrics.} We report four complementary metrics:
\begin{itemize}[leftmargin=*,itemsep=1pt]
  \item \textbf{WER / CER} — standard word- and character-level error
    rates, computed against Gemini~2.5~Pro references.
  \item \textbf{Entity recall} — fraction of ground-truth named entities
    correctly present in the transcript, measured via exact string match
    after case normalisation.
  \item \textbf{Filler recall} — fraction of ground-truth filled pauses
    (\emph{uh}, \emph{um}, \emph{ah}) correctly preserved in the
    verbatim transcript.
\end{itemize}

\noindent\textbf{Baselines.} We compare against five systems spanning the efficiency--quality
spectrum: Parakeet TDT-CTC 110M ~\cite{harper2024parakeet}(the previously deployed production model),
Whisper~\cite{radford2023robust}, Qwen2.5-Omni-3B (un-fine-tuned), AssemblyAI Universal-3-Pro ~\cite{assemblyai2025universal}(a commercial API-based system), and Qwen3-Omni-30B~\cite{xu2025qwen3omni}
(a mixture-of-experts model with 3B active parameters, evaluated
zero-shot to probe the effect of scale without task-specific
adaptation).  All baselines are evaluated on identical test
audio with the same reference transcripts.

\noindent\textbf{Statistical testing.} To assess whether gains from heuristic curation are significant,
we conduct paired bootstrap tests~\cite{koehn2004statistical} (10\,000 iterations)
comparing the entity-heuristic (EH) and random-sample (RS) adapters
on entity recall.  We report the observed difference, 95\% bootstrap
confidence intervals, and $p$-values. 
\begin{table*}[t]   
\centering
\caption{ASR performance across three regions.  Best result per
  metric is \textbf{bold}; second best is \underline{underlined}.
  All metrics computed on the verbatim transcript against
  Gemini~2.5~Pro references.  Qwen3-Omni-30B uses MoE
  (3B active parameters) and is evaluated zero-shot.}
\label{tab:main}
\fontsize{8.5}{10}\selectfont
\begin{tabular}{ll cccc}
\toprule
\textbf{Region} & \textbf{Model}
  & \textbf{WER}$\downarrow$ & \textbf{CER}$\downarrow$
  & \textbf{Entity Recall}$\uparrow$ & \textbf{Filler Recall}$\uparrow$ \\
\midrule
\multirow{7}{*}{\rotatebox{90}{\textbf{India}}}
  & Parakeet          & 13.01 & 7.11 & 53.59 &  0.37 \\
  & Qwen (base)       & 11.02 & 6.28 & 65.34 &  5.33 \\
  & Universal-3-Pro   &  7.22 & 3.58 & \underline{78.64} & 25.37 \\
  & Whisper           &  9.63 & 5.97 & 78.59 &  1.63 \\
  & Qwen3-Omni-30B    &  7.13 & 4.02 & 76.60 & \textbf{91.27} \\
  & Qwen-ft-eh        & \textbf{5.95} & \textbf{3.18} & \textbf{79.52} & 76.79 \\
  & Qwen-ft-rs        & \underline{6.51} & \underline{3.46} & 74.42 & \underline{87.94} \\
\midrule
\multirow{7}{*}{\rotatebox{90}{\textbf{Indonesia}}}
  & Parakeet          & 18.18 & 10.02 & 55.23 &  4.15 \\
  & Qwen (base)       & 13.20 &  7.57 & 70.81 & 16.76 \\
  & Universal-3-Pro   &  9.87 &  5.27 & 80.95 & 32.18 \\
  & Whisper           & 12.70 &  7.69 & 79.16 &  7.12 \\
  & Qwen3-Omni-30B    &  8.43 &  4.74 & \underline{82.18} & \textbf{92.95} \\
  & Qwen-ft-eh        & \textbf{7.36} & \textbf{4.28} & \textbf{84.60} & 85.68 \\
  & Qwen-ft-rs        & \underline{7.67} & \underline{4.36} & 81.73 & \underline{87.14} \\
\midrule
\multirow{7}{*}{\rotatebox{90}{\textbf{LatAm}}}
  & Parakeet          & 21.12 & 12.21 & 54.31 &  1.18 \\
  & Qwen (base)       & 16.76 &  9.85 & 68.79 & 10.36 \\
  & Universal-3-Pro   & 12.96 &  7.75 & 78.63 & 17.61 \\
  & Whisper           & 16.44 & 10.57 & 76.92 &  2.66 \\
  & Qwen3-Omni-30B    & 11.79 &  6.98 & 78.82 & \textbf{82.50} \\
  & Qwen-ft-eh        & \textbf{10.00} & \textbf{5.83} & \textbf{81.87} & 76.47 \\
  & Qwen-ft-rs        & \underline{10.75} & \underline{6.27} & \underline{79.05} & \underline{76.51} \\
\bottomrule
\end{tabular}
\end{table*}

\section{Results}
\label{sec:results}

\subsection{Main Results}
\label{sec:main_results}

Table~\ref{tab:main} presents results for all models across the three
target regions.  We highlight three key findings.

\noindent\textbf{Entity recall.}
The fine-tuned models (qwen-ft-eh) achieve 80--85\% entity recall,
a 26--29~pp improvement over Parakeet (53--55\%).  Gains over stronger
baselines are modest but consistent: qwen-ft-eh exceeds
Universal-3-Pro by up to 4pp, Qwen3-Omni-30B zero-shot by 2--3~pp, and
Whisper by 1--5~pp.  The fine-tuned 3B model matches or exceeds the
30B model despite 10$\times$ fewer total parameters.

\noindent\textbf{WER.}
Qwen-ft-eh reduces WER to 6.0--10.0\%, a 53--60\% relative reduction
over Parakeet.  Relative to the unfine-tuned Qwen2.5-Omni-3B, WER
drops by 40--46\%, confirming the gain stems from adaptation rather
than base model selection.

\noindent\textbf{Filler recall.}
Most baselines discard filled pauses: Parakeet achieves 0.4--4.2\%
and Whisper 1.6--7.1\%.  Qwen3-Omni-30B achieves the highest filler
recall (82.5--93.0\%), likely due to its larger capacity.  Our
fine-tuned models achieve 76--86\%---slightly below Qwen3-Omni-30B but
from a model with 10$\times$ fewer parameters.

\subsection{Effect of Data Curation: EH vs.\ RS Ablation}
\label{sec:ablation}

The central empirical claim of this paper is that heuristic data
curation (EH) produces better entity recall than random sampling (RS),
even when model architecture, training recipe, and data volume are held
constant.  Paired bootstrap tests (10\,000 iterations) confirm
statistically significant gains across all three regions:
India (+4.19~pp, $p < 0.0001$),
Indonesia (+2.84~pp, $p < 0.0001$), and
Latin America (+2.76~pp, $p < 0.0001$),
with 95\% confidence intervals excluding zero.  The effect is largest
for India, possibly because Indian English entity names are more
distinct from the base model's training distribution.

This ablation isolates the contribution of data curation: both EH and
RS use the same base model, LoRA configuration, hyperparameters, and
dual-output format.  The RS adapter itself achieves 74--82\% entity
recall---far above all baselines---making the additional curation gain
all the more meaningful.  EH also achieves equal or better WER than RS
(Table~\ref{tab:main}), confirming that entity-focused curation does
not degrade general transcription quality.

\subsection{Discussion}
\label{sec:discussion}

The gap between WER and entity recall across all baselines underscores
that WER is necessary but insufficient for language-learning
applications: Parakeet achieves 13--21\% WER yet misses nearly half of
all entities, while filler recall below 5\% severely limits fluency assessment. 

Qwen3-Omni-30B achieves comparable entity recall and higher filler
recall \emph{without task-specific training}, but requires 10$\times$
more parameters. Our 3B pipeline demonstrates that data curation paired with LoRA can match or exceed not only a commercial API but also a much larger model. The pipeline also yields consistent gains across all three regions despite distinct phonetic profiles, validating the per-region adapter approach.

\section{Conclusion}
\label{sec:conclusion}

We presented a three-stage pipeline for entity-aware conversational ASR targeting accented English from India, Indonesia, and Latin America. Heuristic SQL filters enrich training data to ${\sim}2.8\times$ the entity density of random sampling; regional LoRA adapters raise entity recall to 80–85\% (from 53–55\%) and filler recall to 76–86\% (from $<$5\%), while reducing WER to 6–10\% at a P95 latency of ${\sim}800$ ms when served via vLLM. Bootstrap tests confirm that curation alone accounts for 2.8–4.2 pp of entity recall gain (p $<$ 0.0001). The fine-tuned 3B model outperforms Whisper and a commercial ASR on entity recall while matching a zero-shot 30B model with 10× fewer parameters. Several limitations qualify these findings. First, test-set references are silver-standard (Gemini 2.5 Pro) rather than fully human-verified, though spotchecks on 250 samples per region show high agreement. Second, there remains a filler recall gap vs. the 30B model suggesting room for dual-output training optimisation. Third, evaluation is limited to English; extending to multilingual conversational settings—particularly code-switching, which is common across all three regions—is an important direction for future work. 

\section{Generative AI Use Disclosure}
Gemini~2.5~Pro was used to generate reference transcripts (Section~3.2).  Claude Sonnet~4.5 was used as the LLM-based
error judge (Section~3.5).  Generative AI tools were used for
editing and polishing the manuscript; all authors reviewed and
take responsibility for the final content.
\bibliographystyle{IEEEtran}
\bibliography{mybib}

\end{document}